\documentclass[conference]{IEEEtran}
\usepackage{cite}

\usepackage{amsmath}
\usepackage{amsfonts}
\usepackage{amssymb}
\usepackage{mathtools}

\ifCLASSINFOpdf
  \usepackage[pdftex]{graphicx}
\else
\fi
\usepackage{algorithmic}
\usepackage{algorithm}
\usepackage{amsmath}
\usepackage{color,soul}
\IEEEoverridecommandlockouts

\begin{document}
%
\title{An Automatic Interaction Detection Hybrid Model for Bankcard Response Classification \thanks{- Support by Atlanticus Services Corporation, Atlanta, GA, USA}}




\author{\IEEEauthorblockN{Yan Wang}
\IEEEauthorblockA{Graduate College\\
Kennesaw State University\\
Kennesaw, Georgia 30144\\
ywang63@students.kennesaw.edu}
\and
\IEEEauthorblockN{Xuelei Sherry Ni}
\IEEEauthorblockA{Department of Statistics and Analytical Sciences\\
Kennesaw State University\\
Kennesaw, Georgia 30144\\
sni@kennesaw.edu}
\and
\IEEEauthorblockN{Brian Stone\\}
\IEEEauthorblockA{Atlanticus Services Corporation\\
Atlanta, Georgia 30348\\
Brian.Stone@atlanticus.com}}


%


\maketitle

\begin{abstract}
Data mining techniques have numerous applications in bankcard response modeling.  Logistic regression has been used as the standard modeling tool in the financial industry because of its almost always desirable performance and its interpretability. 
In this paper, we propose a hybrid bankcard response model, which integrates decision tree based chi-square automatic interaction detection (CHAID) into logistic regression. 
In the first stage of the hybrid model, CHAID analysis is used to detect the possibly potential variable interactions. 
Then in the second stage, these potential interactions are served as the additional input variables in logistic regression.
The motivation of the proposed hybrid model is that adding variable interactions may improve the performance of logistic regression. 
Theoretically, all possible interactions could be added in logistic regression and significant interactions could be identified by feature selection procedures. 
However, even the stepwise selection is very time-consuming when the number of independent variables is large and tends to cause the $p >> n$ problem. 
On the other hand, using CHAID analysis for the detection of variable interactions has the potential to overcome the above-mentioned drawbacks. 
To demonstrate the effectiveness of the proposed hybrid model, it is evaluated on a real credit customer response data set. 
As the results reveal, by identifying potential interactions among independent variables, the proposed hybrid approach outperforms the logistic regression without searching for interactions in terms of classification accuracy, the area under the receiver operating characteristic curve ($ROC$), and Kolmogorov-Smirnov ($KS$) statistics. 
Furthermore, CHAID analysis for interaction detection is much more computationally efficient than the stepwise search mentioned above and some identified interactions are shown to have statistically significant predictive power on the target variable. 
Last but not least, the customer profile created based on the CHAID tree provides a reasonable interpretation of the interactions, which is the required by regulations of the credit industry.  
Hence, this study provides an alternative for handling bankcard classification tasks. \\

\textit{Keywords-decision tree; CHAID; hybrid; logistic regression; bankcard response modeling; credit risk modeling}\\
\end{abstract}


%
\IEEEpeerreviewmaketitle

\section{Introduction}
Recently, financial institutions and banks have been experiencing serious competitions. 
They have extensively started to consider the credit risk and bankcard response of their customers since inappropriate credit decisions may result in huge amount of losses. 
When considering the cases regarding credit card applications or bankcard marketing campaigns, financial institutions usually adopt models to evaluate the applicants or to search strategies to target the consumers. 
Hence, many statistical methods and machine learning tools, including Bayesian probability models \cite{mcneil2007bayesian}, support vector machine and neural networks \cite {huang2004credit}, classification and regression trees \cite{lee2006mining}, have been provided to provision the credit or bankcard scoring developments. \\

After careful review of the bankcard response modeling and credit risk scoring literatures \cite{chen2016financial} \cite{Siddiqi2015credit}, it can be concluded that discriminant analyses and logistic regression are the two widely used techniques in building bankcard response models and credit risk models.
Compared with discriminant analysis, logistic regression has the advantage that it can perform variable regression even if the variable has an abnormal distribution \cite{menard2000coefficients}. {}
Therefore, logistic regression has been acted as a good alternative to discriminant analyses in handling bankcard response problems and credit scoring modeling. \\

Chi-square automatic interaction detection (CHAID) analysis is an algorithm created by Gordon V. Kass in 1980 and it discovers relationships between independent variables and the categorical outcomes \cite{firat2017understanding}. 
Chi-square tests are applied at each of the stages in building the CHAID tree and Bonferroni corrections are usually used to account for the multiple testing that takes place \cite{akin2017use}. 
In general, CHAID analysis can be used for prediction and classification purposes as well as for detection of interactions between variables, such as diseases classification \cite{armstrong2018quantitative}, financial distress prediction \cite{geng2015prediction}, and risk assessment \cite{chen2016financial} \cite{zhou2018deep}.\\

Several studies have deployed the feature selection approaches to produce higher model performances. 
For instance, a combined strategy of feature selection approaches, including linear discriminant analysis, rough set theory, decision tree, and support vector machine classification model was proposed in credit scoring \cite{koutanaei2015hybrid}. 
An evolutionary based feature selection approach was applied in a case study of credit approval data \cite{wang2009evolutionary}.
It is worth to mention that, when implementing feature selection approaches, many research only focuses on the features provided by the original data sets. 
There is limited research that aiming at incorporating variable interactions in the feature selection approach. 
Focusing on identifying variable interactions that could potentially improve the model performance, this study aims at firstly using an efficient method for the detection of variable interactions, then incorporating these variable interactions for feature selections in the following modeling stage. \\

Considering the above-mentioned research, the current study proposed a hybrid model, which integrates decision tree based CHAID analysis into the logistic regression.
The first stage of the hybrid model is the CHAID analysis, aiming at taking advantage of CHAID tree for the detection of potential variable interactions. 
The second stage of the hybrid model is to incorporate these interactions as additional features in logistic regression and the most significant features will be selected by stepwise feature selection approach. 
This proposed hybrid model is supposed to have higher performance compared with the pure logistic regression model, which does not contain any variable interactions. 
The effectiveness and feasibility of the proposed hybrid model is evaluated by using the credit customer response data through cross-validation. 
The performance of the hybrid model is compared with that from the pure logistic model in terms of classification accuracy, area under the curve ($AUC$), and the Kolmogorov-Smirnov ($KS$) test statistics.\\

This paper has been structured as follows. 
Since the bankcard response model is used to demonstrate the effectiveness of the proposed hybrid model, we will firstly review its related work in Section \ref{relatedwork}. In Section \ref{materials}, the experimental materials and methods are presented, including the data description, data pre-processing, CHAID analysis for the detection of interactions, modeling and evaluation. 
The experimental results and discussions are elaborated in Section \ref{resultsanddiscussions}. 
Finally, Section \ref{conclusion} is devoted to the conclusions. \\

\section{Related work} \label{relatedwork}
We will review the literature of commonly used techniques in bankcard response modeling and credit scoring modeling in this section.

\subsection{Logistic Regression}
Logistic regression is a widely used statistical modeling technique which relies on measuring the results with dichotomous outcomes.
As a multivariate method, logistic regression has an automatic regression capacity to analyze many independent variables that have potential relationships with the dependent variable. 
In credit union environment, logistic regression is as efficient and accurate as other techniques such as discriminant analysis and neural networks. 
Furthermore, logistic regression models can determine the conditional probability of a specific observation belonging to a class, given the information of this observation. 
Hence, logistic regression provides a better understanding of the distribution of the financial risk than discriminant analysis \cite{desai1996comparison}. 
As a result, logistic regression has been explored widely in building credit scoring models and bankcard response models. 

\subsection{Chi-square Automatic Interaction Detection}
CHAID analysis is one of the main decision tree techniques and it shapes the result as a tree structure. 
The construction of the tree stops whenever it does not find any significant chi-square value between the dependent variable and the factors. 
Thus, the higher chi-square value nodes come first the tree, whereas, the terminal nodes carry the lowest chi-square value \cite{althuwaynee2014novel}. 
CHAID analysis has been popularly used in many classification and regression studies such as hazard analysis \cite{pradhan2013comparative}, medical research \cite{herschbach2004psychological}, and market segmentation \cite{hsu2007chaid}.
It has also been explored in building credit risk models and has obtained promising results in terms of predictive accuracy and type II errors\cite{ince2009comparison}. \\

Since the resultant CHAID tree framework is based on logical relationship between independent variables, CHAID analysis has also be used for interaction detection in some research \cite{hill1997chi}. 
However, most of the recent research about CHAID analysis focuses on using it as a modeling tool for regression or classification problems rather than a technique for the detection of variable interactions. 


\subsection{Ensemble and Hybrid Models}
Recently, ensemble models, where several learning algorithms being employed to solve one problem, have been applied to improve the model performance \cite{yu2008credit}.
Motivated by the idea of ensemble learning algorithms, many researchers have employed hybrid multistage models or integrated multiple classifiers into an aggregated model to obtain better classification results in credit scoring modeling.  
For instance, in \cite{yu2008credit}, a six-stage neural network hybrid learning approach was proposed and its effectiveness was confirmed using two publicly available credit datasets. 
A two-stage hybrid model using artificial neural networks and multivariate adaptive regression splines was shown to outperform the traditionally utilized discriminant analysis and logistic regression in credit scoring modeling in \cite{lee2005two}. 
In another study, a hybrid approach of the integration of integrate genetic algorithm and dual scoring model was shown to enhance the performance of credit scoring model \cite{chi2012hybrid}.\\

In our proposed hybrid approach, CHAID analysis is applied for the detection of variable interactions instead of the modeling tool in the first stage.
In the second stage, the identified variable interactions are served as additional independent variables into the logistic regression model. 
This hybrid approach, which integrates CHAID analysis into logistic regression, is motived by while different from the above-mentioned approaches. \\

 



\section{Materials and methods} \label{materials}
\begin{figure*}[!t]
\centering
\includegraphics[width=7in]{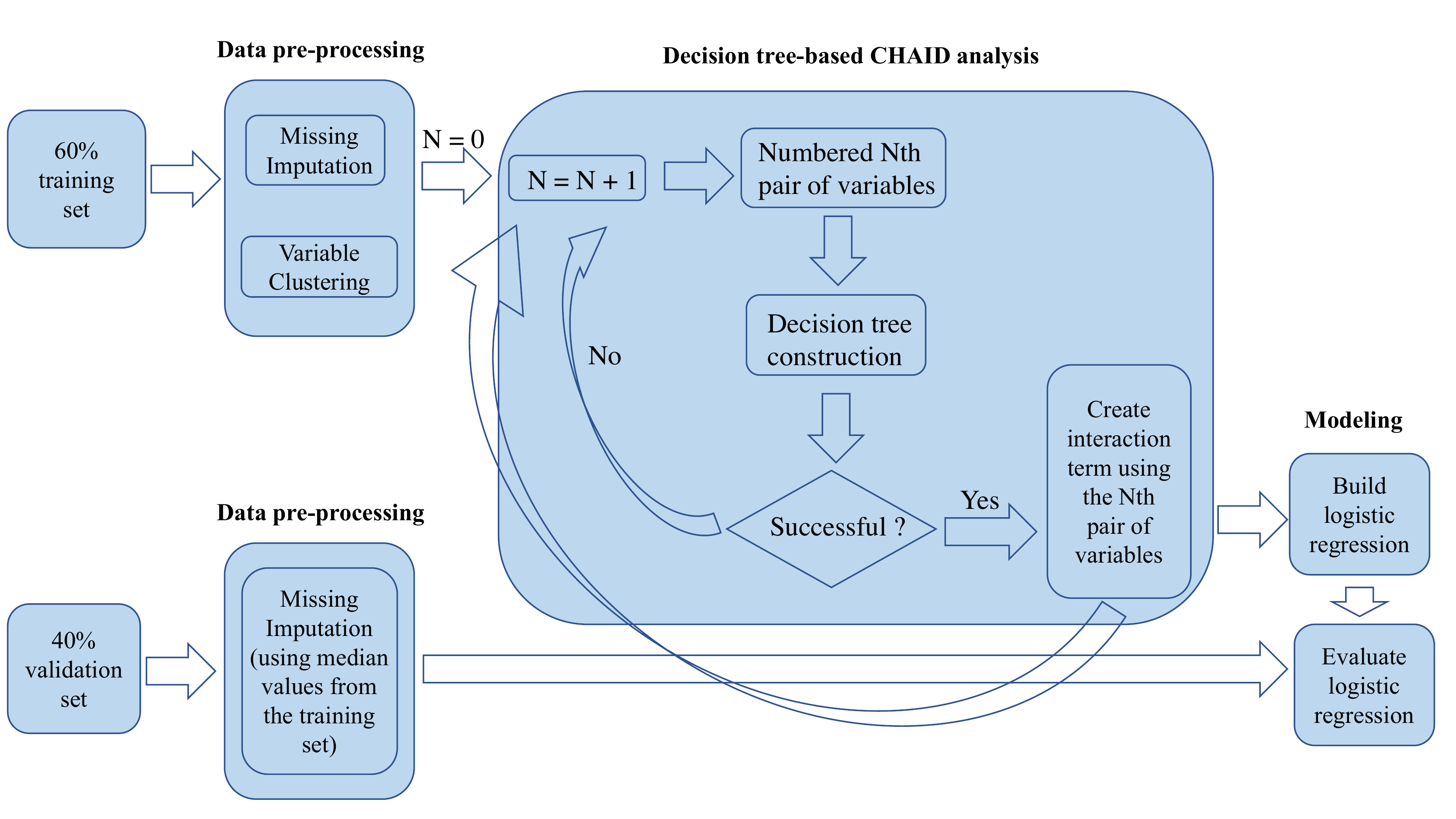}
\caption{An overview of the study procedure. It mainly contains three steps: Data pre-processing, decision tree based CHAID analysis, and modeling and evaluation.}
\label{procedure}
\end{figure*}

In this study, the credit customer response dataset provided by Atlanticus Services Corporation located at Atlanta, GA, USA is used to evaluate the reliability and efficiency of the proposed hybrid decision tree based CHAID and logistic regression model. 
In general, the steps of the study contain data pre-processing, decision tree based CHAID analysis, logistic regression modeling, and model evaluation. 
All the analysis was implemented in SAS 9.4. 
An overview of the study procedure is presented in Fig. \ref{procedure} and details of the study are described as follows.

\subsection{Data Description}
Briefly, the credit customer response dataset was collected from $12,498$ customers representing mainly the customers' credit behaviors. 
$538$ behavior variables were recorded and example factors consist of: customer's number of bankcard accounts, age of the newest account, total balance closed accounts within last three months, total past due amount, and worst status rating reported within one month. 
The target variable $RESP\_FLAG$ has a binary value, where $1$ denotes that customers have the response (have opened the bankcard account) after having the credit card offer while $0$ denotes the opposite. 
Of these $12,498$ customers, around $80\%$ have the responses after having the credit card offer while around $20\%$ show no responses.

\subsection{Data Pre-processing} \label{preprocessing}
Customer records with missing values in the target variable $RESP\_FLAG$ were removed in the very first data pre-processing stage to avoid biased results. 
Then, $40\%$ of the resulting data was set aside for comparison (used as validation data set) and the model was built on the remaining $60\%$ data (training data set). 
During the splitting procedure, stratified sampling was implemented to preserve the original ratio of the outcome in both training and validation data sets. 
As illustrated in the data pre-processing step in Fig. \ref{procedure}, the procedure of missing value imputation was implemented separately for training and validation sets. 
For the training set, variables with more than $90\%$ missing percentage were removed due to the limited information provided. 
Otherwise, median value imputation was implemented to fill the missing values. 
In the meantime, these median values from training set were recorded and were used to impute the missing values on the validation set.\\

To reduce the data dimensionality and decrease the occurrence of multicollinearity problems, hierarchical variable clustering was then applied on the training set and variable with the lowest $1-R^2$\_ratio, as defined in (\ref{1-r2}), in each cluster was selected.  
As a result, $180$ variables were kept after variable clustering, accounting for about $92\%$ variability from the original dataset. 
These variables would be used as the input features in the future modeling stage. 

\begin{equation} \label{1-r2}
1-R^2\_ratio = \frac{1-R^2_{own\_cluster}}{1-R^2_{next\_closest\_cluster}}
\end{equation}\\

\subsection{Complete Stepwise Search for the Detection of Interactions} \label{stepwiseforinterac}
Intuitively, the most straightforward way to look for potential variable interactions is the {\it complete stepwise} search during the modeling procedure. 
That is, consider all the possible pairs of variables as interaction terms, feed them into the model, and use criteria such as $p$-value to filter out the significant interaction terms through stepwise, backward, or forward feature selection approaches \cite{bagherzadeh2016tutorial}. 
Considering that $180$ variables were kept after data pre-processing, there would be $C^{2}_{180}$ combinations of variables. 
As a result, by using (\ref{permutation}), $16,110$ interaction terms would be served as additional independent variables and would go through the feature selection procedure when constructing the models.

\begin{equation}
\label{permutation}
C^k_n = \frac{n!}{(n - k)!k!}
\end{equation}\\

There are many problems by using the above-mentioned complete search for variable interactions. 
After adding all the possible $16,110$ interaction terms into the data set, the resulting number of independent variables ($16,110 + 180 = 16,290$) would be larger than the number of observations (i.e., $12,498$ in this study). 
This will cause ``large p small n problem'' ($p>>n$ problem) and there would be insufficient degrees of freedom to estimate the model coefficients \cite{yin2015sequential}.
Even though no ``large p small n problem'' occurs, the entire processing time for the complete stepwise search would be very long when the data is large. 
In our experiment, we tried to randomly select $10,000$ interaction terms and served them as additional input variables in logistic regression by using SAS $9.4$ on the computer with 3.3 GHz Intel Core I7 processor to estimate the time consuming. 
As a result, it took more than ten hours when either of the stepwise, backward, or forward selection method was applied as the feature selection tool. 
Therefore, it is confident to conclude that for the data used in this study, even no ``large p small n problem'' problem is caused, the complete stepwise search method is not efficient in terms of time consuming. 
Furthermore, this complete stepwise search methodology tends to cause multicollinearity problem and thus could offset the advantages brought from the hierarchical variable clustering step illustrated in Fig. \ref{procedure}. 
Therefore, a more efficient way for the detection of the possible variable interactions is needed and this motivates the occurrence of the proposed hybrid model. \\

\subsection{CHAID Analysis for the Detection of Interactions} \label{CHAIDinteraction}
In the proposed hybrid decision tree based CHAID and logistic regression model, CHAID analysis was firstly implemented to identify the pairs of variables with possibly potential interactions. 
In particular, we propose to use the CHAID decision tree idea to identify potential significant interactions.  In a CHAID tree, a predictor and its best segmentation is identified to split the node (or grow the tree) based on an adjusted significance test according to the chi-square statistic.  The tree will keep growing if a significant chi-square statistic can be found.   When applying this idea in looking for potential interactions, we only feed a pair of variables in to the CHAID decision tree.  If the CHAID tree was successfully built (Fig. \ref{tree}), we consider that there exist potential interactions in the current pair of variables. 
That is, the effect of one variable on the target variable $RESP\_FLAG$ depends on its partner. 
Therefore, the interaction term will be created and it will enter the logistic regression modeling by using stepwise method for feature selection. 
Otherwise, no interactions exist for the current pair and the iteration will continue using the next pair of variables.\\

Please note that it is possible the CHAID tree is successfully built using only one variable from the pair. 
In this case, the interaction term would still be created to avoid any missing of the potential interactions. 
If this newly created interaction term does not provide useful information in the prediction of target variables, it would be removed in the stepwise feature selection procedure in the following logistic regression modeling stage. 
Furthermore, during the modeling stage, variance inflation factor ($VIF$) values for the variables would be checked to avoid the entering of redundant interactions in the final model. \\

Pre-setting of the criteria of CHAID analysis is essential, because that will positively or negatively affect the tree size, and more importantly, the processing time \cite{ivanvcevic2016decision}. 
To find the optimal values of these criteria settings, we did various experiments that include a series values for the criteria setting. 
The experiments were established on the computer with 3.3 GHz Intel Core I7 processor. 
With the purpose of avoiding too long processing time (we limit the time for the CHAID analysis being within five hours in this study), avoiding too complex decision tree structures, including more potentially significant interactions, and excluding possibly useless interactions, we finally set the criteria of CHAID analysis as follows:
$p$ value for the chi-square test, which was used to control merging or creating a new branch, was set to $0.3$. 
In order to obtain more nodes in the tree, we set the minimum node size for split to 18 and the minimum leaf size to 10. 
For each node, the maximum number of branches was set to 3 to avoid too large tree structures. 
The maximum depth of the tree was set to 15 through experiments. 
This value could avoid too long processing time as well as too complex tree structure. \\

\subsection{Properties of CHAID Analysis Used} \label{CHAIDproperty}
It is worth mentioning that different from many studies that use decision tree based CHAID for classification, in this study, CHAID is used to identify potential variable interactions. 
Comparing with other decision tree methods such as Classification and Regression Tree (CRT) and Quick, Unbiased, Efficient Statistic Tree (QUEST), CHAID method has the major properties as follows, which is the main reason why this method was used for the interaction detection in this study:

\begin{itemize}
\item CRT and QUEST are binary trees and are not able to produce multi-branches based tree. 
In contrast, CHAID builds non-binary tree containing two or more branches growing from a single node \cite{wilkinson1992tree}. 
This is helpful to identify the complex interactions among variables. \\

\item Tree pruning tasks are usually used in CRT methods, whereas in the case of CHAID method they are not required \cite{loh2011classification}. 
This could largely decrease the computational time, especially when the data set is relatively large. \\

\item CHAID can model both categorical or ordinal data. 
The continuous data is automatically converted to ordinal during the analysis \cite{diaz2016chaid}.
Since most of the variables used in this study are continuous, CHAID method is the first choice. \\

\item Data summarizing performance in CHAID analysis is equivalent to stepwise regression models such as logistic regression \cite{althuwaynee2014novel}. 
However, the customer profile based on CHAID tree could be created naturally and a better interpretation about variable interactions could be provided (see Table \ref{profile}). 
This makes the CHIAD method being preferable than other interaction detection methodologies in regulated industries.\\
\end{itemize}


\subsection{Modeling -- the Hybrid Model and the Pure Logistic Model}
The logistic regression aims at measuring the results with dichotomous variables such as $1$ and $0$. 
It builds a statistical model to predict the logit transformation of the occurrence probability of the target variable $RESP\_FLAG$ in this study. 
The format of logistic regression can be represented in (\ref{logistic}), where, $p$ denotes the probability of the occurrence of $RESP\_FLAG$, $n$ is the number of independent variables, and $\beta_{i}$ are the coefficients of the independent variables $x_{i}$. 

\begin{equation}
\label{logistic}
p = \frac{1}{1+e^{-(\beta_{0} + \beta_{1}*x_{1} + ... + \beta_{n}*x_{n})}}
\end{equation}\\

In this study, to show the effectiveness of the interaction terms identified by CHAID analysis, two models were built and compared during the modeling procedure:

\begin{itemize}
\item The proposed hybrid decision tree based CHAID analysis and logistic regression model, denoted as the hybrid model in this paper, was built following the steps illustrated in Fig. \ref{procedure}. 
That is, after CHAID analysis, newly created interaction terms were used as additional independent variables and a stepwise feature selection approach was used to select the most important contributed predictors for the target variable $RESP\_FLAG$ in the logistic regression.
The significant levels of the entering and leaving the model for the variables were set to 0.15. \\

\item The logistic regression model without CHAID analysis, denoted as the pure logistic model in this paper, was built following the steps except the decision tree based CHAID analysis stage illustrated in Fig. \ref{procedure}. 
That is, no interaction terms were created while the same stepwise feature selection approach was used in logistic regression.\\ 
\end{itemize}

By comparing the performances of the two models, the purpose is to demonstrate the effectiveness of the newly created interactions through CHAID analysis as well as to show the superiority of the proposed hybrid model.\\

\subsection{Model Evaluation}
The models were firstly evaluated using receiver operating characteristic (ROC) method. 
Since $AUC$ has been the most common measure of discrimination for prediction models with binary outcome, we take advantage of $AUC$ for its popularity in this study \cite{hanley1982meaning}.
Both training and validation sets were used to measure $AUC$ for comparing the performances of the proposed hybrid model and the pure logistic regression model when different number of variables are kept. 
When keeping the same number of variables, the desirable model should have higher $AUC$ value. \\

The second evaluation measure used in this study is the classification accuracy. 
After the observations are classified as a binary response into (1,0) categories (using cutoff valued 0.5, i.e., if observations have predicted value larger than 0.5 in logistic regression, they are classified to category 1 while those with predicted value no larger than 0.5 belong to category 0), four possible consequences named true positive (TP), true negative (TN), false positive (FP), and false negative (FN) are produced. 
The classification accuracy can be obtained using (\ref{accuracy}) and higher accuracy is expected to be from the better model.

\begin{equation}
\label{accuracy}
accuracy = \frac{TP+TN}{TP+TN+FP+FN}
\end{equation}\\

The last evaluation measure applied is the $KS$ statistic, which quantifies a distance between the empirical distribution function of two samples. 
The $KS$ statistic $D_n$ is defined in (\ref{KS}), where $F_n(x)$ and $F_p(x)$ denotes the cumulative density function of the classifier scores for negatives and positives, respectively \cite{justel1997multivariate}. 
In general, larger $KS$ statistic value denotes the better goodness of fit of the model. 

\begin{equation}
\label{KS}
D_n = \max_{x} |F_n(x) - F_p(x)|
\end{equation}\\

\section{Results and discussions} \label{resultsanddiscussions}
\subsection{Result of CHAID Analysis for Interaction Detections}
\begin{figure}[!t]
\centering
\includegraphics[width=3.4in]{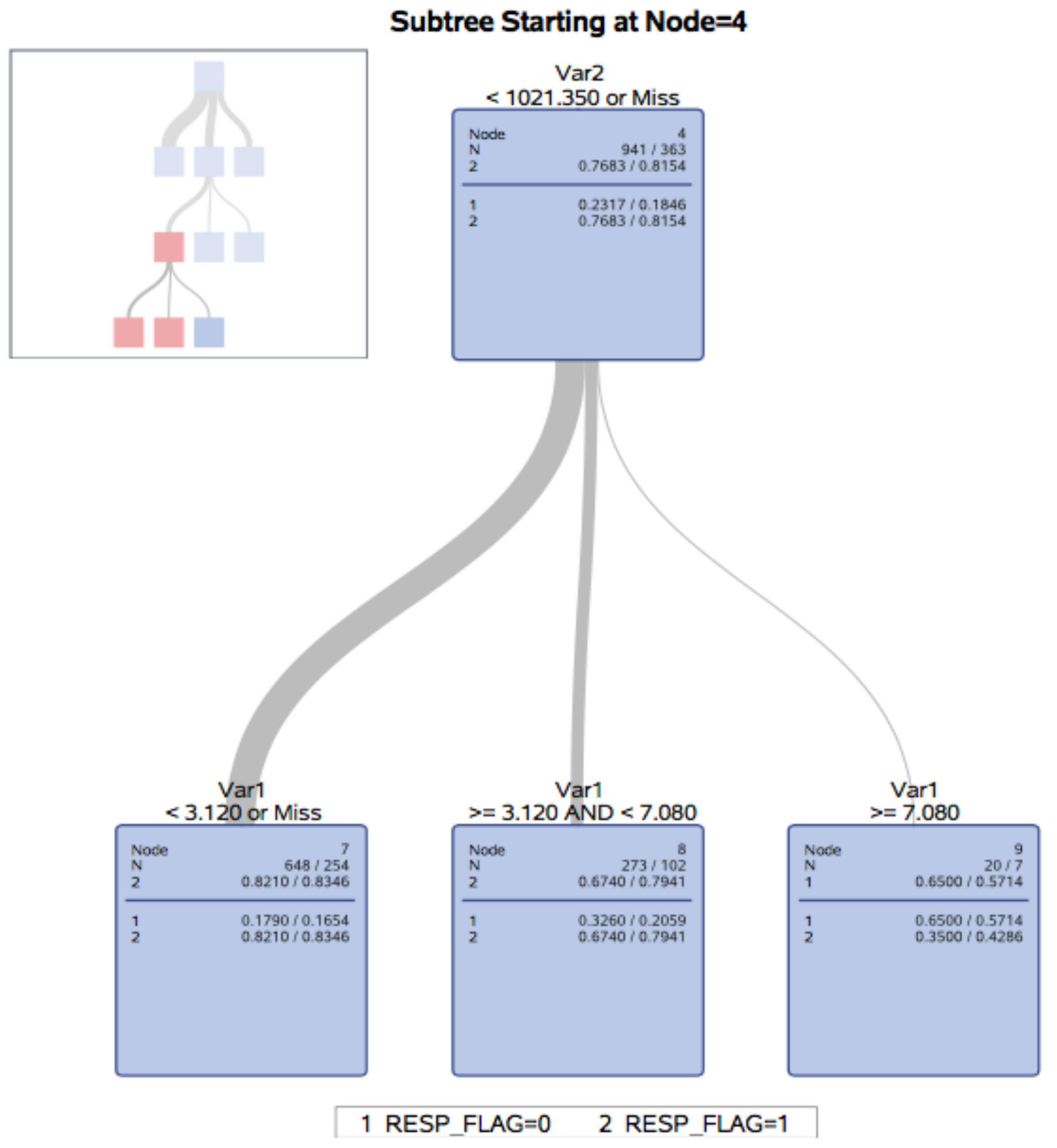}
\caption{An illustrative example of the CHAID subtree. The entire structure of the CHAID tree is represented on the top left corner and the structure of the subtree starts from node 4 for illustrative purpose. }
\label{tree}
\end{figure}

As stated in Section \ref{CHAIDinteraction}, during each iteration in CHIAD analysis, one of the $16,110$ pairs of variables was used to build the CHAID tree. 
To illustrate the result of the analysis, take a certain pair of variables ($Var1$, $Var2$) as an example. 
Fig. \ref{tree} shows the subtree structure starting from node 4 after CHAID analysis on $Var1$ and $Var2$. 
Since the decision tree is successfully built, it is reasonable to assume that the effect of $Var1$ on the target variable $RESP\_FLAG$ depends on $Var2$.
Therefore, the interaction term $Var1*Var2$ is created and would be served as an additional input into the following logistic regression modeling stage. 
As the final result of CHIAD analysis, $1,025$ pairs of variables are shown to be successful in the tree construction. 
Therefore, $1,025$ interaction terms are finally created and their predictive power on the target variable $RESP\_FLAG$ would be determined using the stepwise feature selection procedure in logistic regression. 
It is worth to mention that, the entire CHAID analysis took only about two hours, which is much shorter than the ten hours needed for the interaction search by using either of the stepwise, backward, or forward selection method in logistic regression mentioned in \ref{CHAIDinteraction}. 
Therefore, CHAID analysis could save at least 80\% of the processing time and is much more computationally efficient based on the dataset used in this paper.\\

\subsection{Result of the Hybrid Model and the Pure Logistic Model}
Tables \ref{emsembleresult} and \ref{logisticresult} show the classification accuracy and $AUC$ by using different number of variables on both training and validation sets produced by the proposed hybrid model and the pure logistic model, respectively. 
When decreasing the number of variables, the Wald chi-square values in logistic regression were referred to and variables with lowest Wald chi-square value were removed first. 
Furthermore, a multicollinearity test was performed to calculate the $VIF$ values of the variables for each model. 
Test shows that all the variables selected by the models in Tables \ref{emsembleresult} and \ref{logisticresult} are not interdependent ($VIF$ $< $ 10). \\

Since for each of the models in Tables \ref{emsembleresult} and \ref{logisticresult}, classification accuracy and $AUC$ are very similar on the training and validation sets, there is no evidence for the occurrence of over-fitting problem.  
As expected, when the number of selected variables decreases, the classification accuracy and $AUC$ generally demonstrate a decreasing trend in both training and validation sets for both proposed model and the pure logistic model. 
Most importantly, it is observed that when the same number of selected variables was used, the hybrid model generally outperforms the pure logistic model in terms of classification accuracy and $AUC$ on both training and validation sets.

\begin{table}[!t]
\renewcommand{\arraystretch}{1.3}
\caption{Result of the hybrid model}
\label{emsembleresult}
\centering
\begin{tabular}{|p{13mm}||p{13mm}|p{14mm}|p{13mm}|p{14mm}|p{14mm}}
\hline
Number of \newline selected variables & Accuracy on \newline train(\%) & Accuracy on \newline validation(\%) & $AUC$ on \newline train(\%) & $AUC$ on \newline validation(\%)\\
\hline
30 & 83.86 & 82.40 & 83.45 & 79.46\\
27 & 83.81 & 82.26 & 83.17 & 79.21\\
24 & 83.48 & 81.86 & 82.56 & 78.85\\
21 & 83.14 & 81.64 & 82.22 & 78.49\\
18 & 82.97 & 81.70 & 81.58 & 77.90\\
15 & 82.86 & 81.30 & 80.59 & 77.08\\
12 & 82.45 & 80.90 & 79.87 & 76.32\\
\hline
\end{tabular}
\end{table}

\begin{table}[!t]
\renewcommand{\arraystretch}{1.3}
\caption{Result of the pure logistic model}
\label{logisticresult}
\centering
\begin{tabular}{|p{13mm}||p{13mm}|p{14mm}|p{13mm}|p{14mm}|p{14mm}}
\hline
Number of \newline selected variables & Accuracy on \newline train(\%) & Accuracy on \newline validation(\%) & $AUC$ on \newline train(\%) & $AUC$ on \newline validation(\%)\\
\hline
30 & 83.73 & 81.56 & 82.48 & 77.59\\
27 & 83.56 & 81.78 & 82.18 & 77.20\\
24 & 82.87 & 80.92 & 80.80 & 76.17\\
21 & 82.58 & 80.82 & 80.34 & 76.16\\
18 & 82.36 & 80.80 & 79.69 & 75.70\\
15 & 81.90 & 80.86 & 78.99 & 75.76\\
12 & 81.73 & 80.70 & 78.53 & 75.63\\
\hline
\end{tabular}
\end{table}

\subsection{Model Comparison}

Fig. \ref{KS}, where $y$-axis represents the $KS$ statistic and $x$-axis represents the number of selected variables, shows the $KS$ statistics in the hybrid model and pure logistic regression. 
As can be noticed, the $KS$ statistics does not change too much on validation set with the changing the number of selected variables on both models. 
However, the hybrid model always produce a higher $KS$ statistics than the pure logistic model in both training and validation sets when the number of selected variables is fixed. 
Therefore, by considering all the results described earlier, it is confirmed that the proposed hybrid model outperforms the results using logistic regression without interactions.\\

\begin{figure}[!t]
\centering
\includegraphics[width=3.4in]{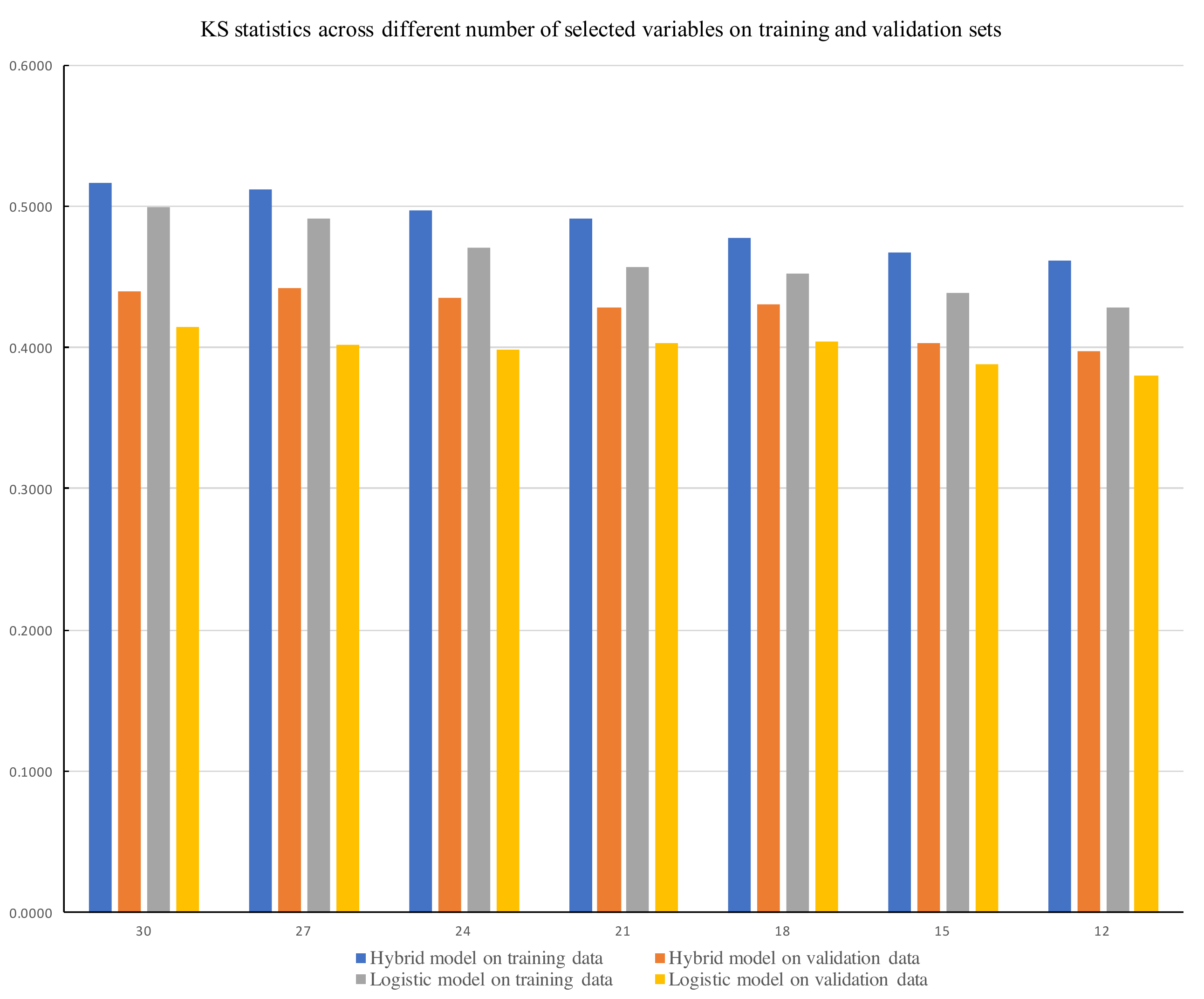}
\caption{Results of $KS$ statistics of hybrid model and the pure logistic model.}
\label{KS}
\end{figure}

Another important issue to consider is that, it is time and economy consuming in collecting customers' information in the credit research domain, hence a practical while reliable bankcard response model should not contain too many independent variables. 
Using the criterion that at least $KS$ statistics valued $0.4$ should be reached on the validation set, the hybrid model with $15$ selected variables is considered to be the best model in this study. 
To demonstrate the effectiveness of the identified interaction terms in the best model, descriptions of the selected variables are summarized in Table \ref{varlist}.
It is worth to mention that, all the variables listed in Table \ref{varlist} have $p$ values less than $0.001$ from the Wald chi-square test and $VIF$ values less than $10$. 
Furthermore, according to Table \ref{varlist}, there are 4 interaction terms entering the best model, indicating the necessity of interaction detections. 
Therefore, these statistically significant interaction terms in the hybrid model further confirmed that the proposed hybrid approach, outperforms the pure logistic regression model and hence provides an alternative in handling bankcard response classifications.

\begin{table}[!t]
\renewcommand{\arraystretch}{1.3}
\caption{List of selected variables in the best model}
\label{varlist}
\centering
\begin{tabular}{|p{11mm}||p{65mm}|}
\hline
Variable & Description \\
\hline
$x_1$ & Number of inquiries within 1 month\\
$x_2$ & Percent balance to high credit open department store accounts with update within 3 months\\
$x_3$ & Age of the newest bankcard account\\
$x_4$ & Age of newest judgment public record item\\
$x_5$ & Dismissed bankruptcy public record within 24 months\\
$x_6$ & Total loan amount open mortgage accounts with update within 3 months\\
$x_7$ & Total balance open student loan accounts with update within 3 months\\
$x_8$ & Age of newest data last activity installment accounts paid as agreed\\
$x_9$ & Total balance closed bankcard accounts with update within 3 months\\
$x_{10}$ & Total past due amount installment accounts\\
$x_{11}$* $x_{12}$& Number open bankcard accounts with update within 3 months with balance $\ge$ 75\% loan amount * Total balance open retail accounts with update within 3 months\\
$x_{13}$*$x_{14}$ & Number department store accounts worst rating 120 to 180 or more days past due within 6 months or major derogatory event within 24 months * Number installment accounts opened within 6 months\\
$x_{15}$ & Age of newest mortgage account\\
$x_{11}$*$x_{16}$ & Number open bankcard accounts with update within 3 months with balance $\ge$ 75\% loan amount * Accounts worst rating ever 90 days past due\\
$x_{11}$*$x_{17}$ & Number open bankcard accounts with update within 3 months with balance $\ge$ 75\% loan amount * Percent utility inquiries within 3 months to inquiries within 24 months\\
\hline
\end{tabular}
\end{table}

\subsection{Customer Profile}
One thing the financial industries concerned when building the bankcard response model or the credit risk model is the customer profile. 
After detecting the existence of the interactions of the variables, one more important issue is to understand how these variables interact. 
Take the interaction $x_{11}$*$x_{16}$ term shown in Table \ref{varlist} as an illustrative example. 
Since this interaction term is statistically significant in the proposed hybrid model, we are more concerned about the customer response rate (i.e., percentage of $RESP\_FLAG = 1$) profile when considering these two variables alone. 
By taking advantage of the CHAID tree produced by the CHAID analysis in the hybrid model, the customer response rate profile could be created in Table \ref{profile} and a better interpretation on the interactions between $x_{11}$*$x_{16}$ could be provided. 
As Table \ref{profile} indicates, customers who have at least two bankcard accounts with update within 3 months with balance $\ge$ $75\%$ loan amount (i.e., $x_{11} \ge 2$) present a lower response rate than those customers who have only one or even zero such bankcard accounts (i.e., $x_{11} < 2$). 
Furthermore, for customers with $x_{11}$ valued less than 2, the response rate does not depend on another variable $x_{16}$. 
However, on the other hand, for customers with $x_{11}$ valued at least 2, the response rate does depend on another variable $x_{16}$. 
As the result shown in Table \ref{profile}, increasing $x_{16}$ values indicates lower response rate. 
Therefore, the created customer profile could provide the financial institutions a thorough understanding about the behaviors of their customers. 

\begin{table}[!h]
\renewcommand{\arraystretch}{1.3}
\caption{Customer Response Rate Profile by $x_{11}$ and $x_{16}$}
\label{profile}
\centering
\begin{tabular}{|p{14mm}||p{30mm}|p{24mm}|}
\hline
$x_{11}$ & $x_{16}$ & Response Rate (\%)\\
\hline
$<$ 2 & any valid value & 84.69\\
$\ge$ 2 & $<$ 557.10 & 79.40\\
$\ge$ 2 & $\ge$ 557.10 and  $<$ 1021.35 & 71.07\\
$\ge$ 2 & $\ge$ 1021.35 & 54.69\\
\hline
\end{tabular}
\end{table}

\section{Conclusion} \label{conclusion}
The bankcard response models play an important role in helping financial companies in their decision making. 
Logistic regression is one of the popularly utilized techniques in the credit research domain. 
This technique focus on exploring linear relationships among variables, especially among independent and dependent variables. 
In credit card research area, there may exist complex interactions among independent variables, that is, the relationship between one predictor and the target variable, depends on the value of another independent variable. 
Therefore, adding interaction terms during the modeling procedure has the potential to produce better model performances. 
However, the possible number of interactions increases dramatically as the number of independent variables increases, which could largely increase the computational time. 
In this situation, an efficient way for interaction detection is needed.\\

The main objective of this research is to propose a hybrid data mining approach, which integrates decision tree based CHAID analysis into logistic regression model to improve the performance for bankcard response classification. 
The rationale underlying the analyses is firstly using the decision tree based CHAID method, a novel multivariate tool, to identify the potential interactions among independent variables. 
Then these newly created interactions are served as additional independent variables in the logistic regression for further feature selection through stepwise procedure.\\ 

The effectiveness of the proposed hybrid model is demonstrated by using the credit customer response dataset provided by Atlanticus Services Corporation located at Atlanta, GA, USA. 
The proposed hybrid model and the pure logistic regression model (without CHAID analysis) were evaluated and compared when implementing credit response classification tasks.  
It is shown that by identifying variable interactions using CHAID method, the hybrid model outperforms the pure logistic regression model in terms of classification accuracy, $AUC$, and $KS$ statistics. 
By selecting the model with 15 variables, it is found that 4 of these 15 variables are the interaction terms identified by CHAID analysis and they are all statistically significant. 
This could further confirm the necessity of the detection of variable interactions for predicting the outcomes. 
Furthermore, CHAID method is more computationally efficient in identifying potential interactions when compared with adding all possible interactions in either of the stepwise, backward, or forward feature selection procedures in logistic regression. 
Also, the customer profile created based on the CHAID tree could provide a better understanding about the variable interactions. \\

As a general conclusion, the advantages of the proposed hybrid decision tree based CHAID analysis and logistic regression model in the current research are:  
\begin{itemize}
  \item Most of the recent research uses CHAID analysis as a prediction or classification technique. Different from this, powerful decision tree based CHAID analysis is applied to identify potential variable interactions, which has not been used widely in bankcard response modeling. \\

  \item More importantly, CHAID analysis for interaction detections outperforms the complete stepwise searching for significant interactions by either of the stepwise, backward, or forward feature selection methods in logistic regression in terms of its processing time.
  Based on the dataset used in this study, CHAID analysis for interaction detections could save at least 80\% of the running time compared with the complete stepwise searching for significant interactions in logistic regression. 
  \\

  \item Furthermore, some of the interaction terms identified by CHAID analysis are shown to be statistically significant in the proposed hybrid model and hence enhance the model performance compared with pure logistic regression without interaction terms. \\

   \item Finally, by taking advantage of the CHAID tree produced by CHAID analysis, the customer profiles could be created and a better interpretation about the variable interactions could be provided.\\
\end{itemize}

Therefore, the proposed hybrid model in this study offers a valuable aid for financial industries in handling bankcard response classifications or credit scoring tasks. \\

\section*{Acknowledgment}
The authors would like to thank Atlanticus Services Corporation (located at Atlanta, GA, USA) for providing the customer credit response data set. \\



%
\bibliographystyle{IEEEtran}
\bibliography{CHAID}




\end{document}